\documentclass[10pt,twocolumn,letterpaper]{article}

\usepackage{cvpr}      
\definecolor{cvprblue}{rgb}{0.21,0.49,0.74}
\usepackage[pagebackref,breaklinks,colorlinks,allcolors=cvprblue]{hyperref}

\usepackage{multirow}
\usepackage{graphicx}
\usepackage{amsmath}
\usepackage{amssymb}
\usepackage{booktabs}

\usepackage{tabularx}
\usepackage{array}
\usepackage{ragged2e}

\usepackage{orcidlink}
\usepackage{parskip}
\usepackage[capitalize]{cleveref}
\crefname{section}{Sec.}{Secs.}
\Crefname{section}{Section}{Sections}
\Crefname{table}{Table}{Tables}
\crefname{table}{Tab.}{Tabs.}

\def\paperID{*****} 
\def\confName{CVPR}
\def\confYear{2026}

\title{Applications of Large Models in Medicine}
\begin{document}
\author{
    YunHe Su\textsuperscript{1}, Zhengyang Lu\textsuperscript{2†}, Junhui Liu\textsuperscript{3†},\\
    Ke Pang\textsuperscript{4}, Haoran Dai\textsuperscript{5}, Sa Liu\textsuperscript{6},\\
    Yuxin Jia\textsuperscript{7}, Lujia Ge\textsuperscript{8}, Jing-min Yang\textsuperscript{*}
}

\maketitle

\noindent
\textsuperscript{1} The First Clinical Medical School, MuDanJiang Medical University, Heilongjiang, China\\
{\tt\small Yunhe.Su@outlook.com} \orcidlink{0009-0007-9051-5620} \\
\textsuperscript{2} School of Design, Jiangnan University, Wuxi, China\\
{\tt\small luzhengyang@jiangnan.edu.cn} \orcidlink{0000-0002-1540-0678} \\
\textsuperscript{3} Kunming, Yunnan, China\\
{\tt\small Chinafengwu1025@gmail.com} \\
\textsuperscript{4} Department of Anesthesiology, Third Xiangya Hospital, Central South University, Changsha, Hunan, China\\
{\tt\small pangke97@gmail.com} \\
\textsuperscript{5} Nanjing Medical University, Nanjing, China\\
{\tt\small 2497006106@qq.com} \\
\textsuperscript{6} University of California, Davis, California, USA\\
{\tt\small liusalisa6363@gmail.com} \orcidlink{0000-0002-7249-6020} \\
\textsuperscript{7} Brown School, Washington University in St. Louis, Louis, Missouri, USA\\
{\tt\small j.yuxin@wustl.edu} \\
\textsuperscript{8} School of Basic Medical Sciences, Capital Medical University, Beijing, China\\
{\tt\small gelujiagelujia@qq.com} \orcidlink{0009-0009-6970-1888} \\

Correspondence: Jing-min Yang, WestChina Biomedical Big Data Center, WestChina Hospital, Sichuan University, Chengdu, China\\
{\tt\small yangjingmin2021@163.com}

\vspace{0.2cm}
\noindent
\textit{† These authors contributed equally to this work.}

\vspace{0.5cm}
\begin{abstract}{This paper explores the advancements and 
applications of large-scale models in the medical field, with a 
particular focus on Medical Large Models (MedLMs). These models, 
encompassing Large Language Models (LLMs), Vision Models, 3D 
Large Models, and Multimodal Models, are revolutionizing 
healthcare by enhancing disease prediction, diagnostic 
assistance, personalized treatment planning, and drug discovery. 
The integration of graph neural networks in medical knowledge 
graphs and drug discovery highlights the potential of Large 
Graph Models (LGMs) in understanding complex biomedical 
relationships. The study also emphasizes the transformative role 
of Vision-Language Models (VLMs) and 3D Large Models in medical 
image analysis, anatomical modeling, and prosthetic design. 
Despite the challenges, these technologies are setting new 
benchmarks in medical innovation, improving diagnostic accuracy, 
and paving the way for personalized healthcare solutions. This 
paper aims to provide a comprehensive overview of the current 
state and future directions of large models in medicine, 
underscoring their significance in advancing global health.}
\end{abstract}
\\ \\
\noindent \textbf{Keywords:} Medical Models, Large Language Models, Vision Models, Multimodal Models, Drug Discovery

\begin{figure*}[!ht]
    \centering
    \includegraphics[width=0.98\textwidth]{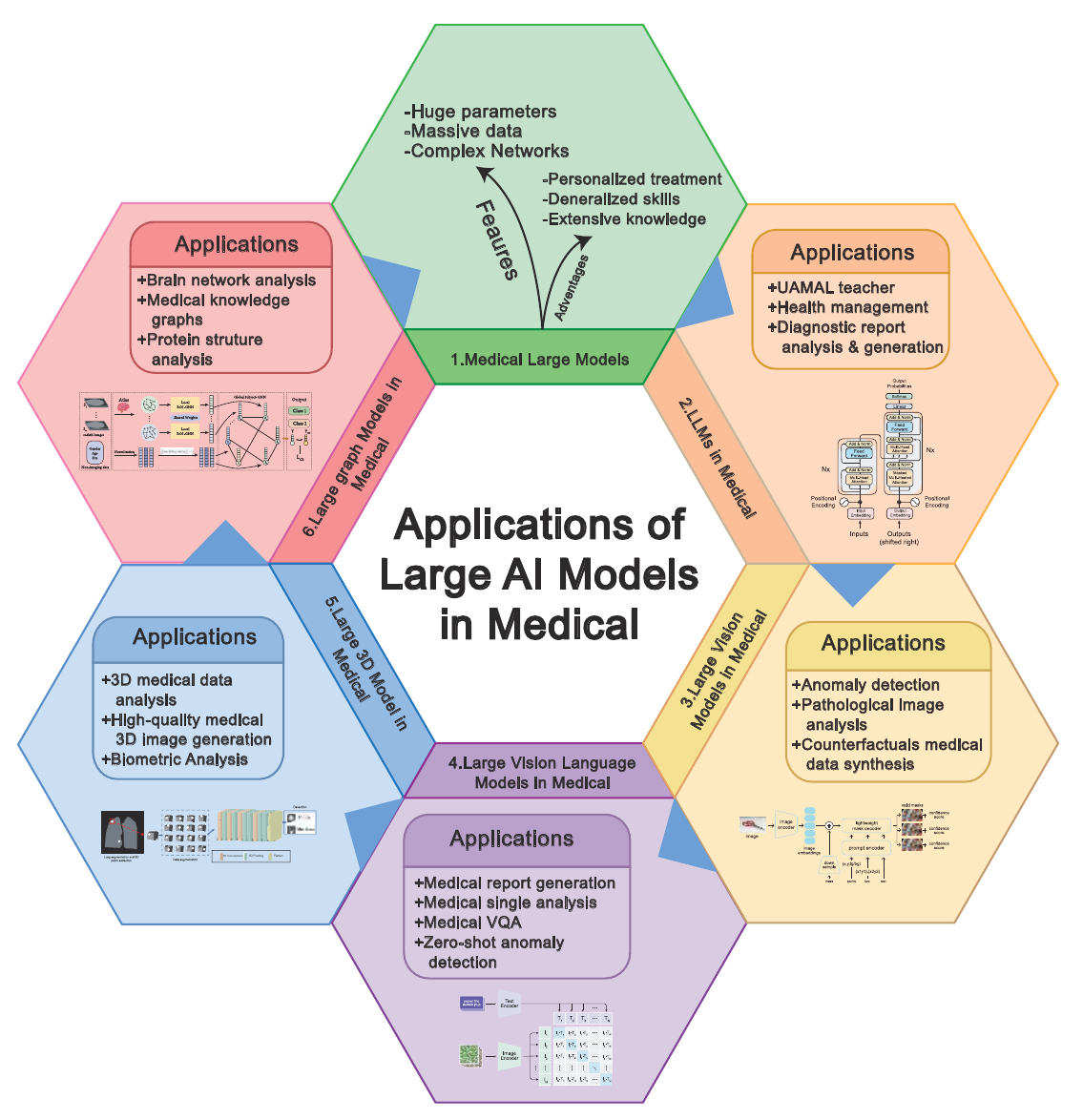}
    \vspace{-10pt}
    \caption{The overall structure of the survey.}
    \label{fig:exp_emp_construct}
\end{figure*}
\section{Introduction}
\label{sec:intro}

Medical Large Models (MedLMs) refer to a class of large-scale 
artificial intelligence models specifically trained to handle 
and analyze various types of medical-related data, such as 
clinical text\cite{Rajkomar2019}, imaging data and genetic 
information. They are typically based on deep learning and 
neural network technologies, enabling them to perform a variety 
of tasks in the medical field, including disease prediction, 
diagnostic assistance, personalized treatment planning, and drug 
development. The core advantage of MedLMs lies in their powerful 
data processing capabilities and their ability to learn from 
vast amounts of data.

MedLMs can be categorized into distinct types, each serving 
different applications based on the data they handle. Large 
Language Models (LLMs) are designed primarily for processing 
clinical textual data (like MedPaLM), such as electronic health 
records (EHRs) such as \cite{zhu2021xanthomatous,
zhu2020suprasellar,zhu2020long,li2023pediatric}. These models 
excel at extracting pertinent information from a wide range of 
medical texts, including patient histories, symptoms, and 
treatment instructions. This capability reduces the manual 
workload for healthcare professionals and provides crucial 
decision-making support. Furthermore, LLMs are instrumental in 
generating clinical pathways, assisting in the creation of 
personalized treatment plans by analyzing a vast corpus of 
medical literature and integrating the latest clinical 
guidelines \cite{Xiao2024} with practical clinical validations 
\cite{PARK2023218,boussina2024large}.

Another major category is Vision Models, which primarily deal 
with medical imaging data. These models, typically based on 
convolutional neural networks (CNNs), have proven effective in 
tasks such as detecting early-stage cancers through the analysis 
of medical images. For example, CNNs have been applied to detect 
skin cancer with dermatologist-level accuracy and are also used 
in the detection of lung and breast cancers, providing 
physicians with fast and reliable diagnostic support. The 
ability to detect even minute abnormalities in images, with 
accuracy comparable to that of experienced specialists, has made 
vision models a powerful tool in modern medical practice\cite
{Esteva2017}.

3D Large Models, which focus on volumetric data analysis, 
represent another critical category within MedLMs. These models 
utilize 3D convolutional neural networks (CNNs) to handle 
medical images in three-dimensional formats, such as CT or MRI 
scans. By accurately segmenting tumors and analyzing their 
spatial characteristics, 3D models assist in determining tumor 
locations and volumes, essential for surgical planning. 
Additionally, these models contribute to virtual surgery 
simulations, allowing medical practitioners to plan surgeries 
more effectively and mitigate potential risks . The integration 
of 3D models into clinical practice is enhancing both the 
precision of tumor detection and the planning of surgical 
interventions \cite{Zhou2023}. 

In addition to these single-modal models, Multimodal Models 
integrate multiple types of data, such as clinical text, 
imaging, and genomic data, to provide a more comprehensive 
understanding of a patient's condition. By combining diverse 
data sources, these models enhance diagnostic accuracy and the 
ability to develop personalized treatment plans. For instance, 
multimodal models have been successfully used to improve the 
early diagnosis of lung cancer by combining CT images with 
clinical records, thus providing more accurate diagnostic 
insights . These models also play a crucial role in 
personalizing treatment plans for complex conditions, such as 
breast cancer, by integrating genomic data, imaging, and 
clinical histories\cite{Yang2020}.

Graph Large Models (Graph Neural Networks, GNNs) are another key 
type of MedLM, particularly in the field of genomics. GNNs are 
designed to analyze the relationships between genes, diseases, 
and therapeutic targets. By studying gene interaction networks, 
GNNs can predict disease risk factors and identify potential 
biomarkers, offering novel insights into early diagnosis and 
potential treatment options. Graph Large Models leveraging 
transformers with large parameters and trained with  large 
biomedical datasets. These models have shown great promise in 
areas such as cancer risk prediction, where they analyze the 
complex interactions between genes and disease.

Medical large models (MedLMs) are bringing new possibilities to 
healthcare. They offer applications in disease prediction, 
diagnostics, treatment planning, and drug development. These 
models analyze extensive medical data and identify patterns that 
may indicate disease risks. For example, genetic interaction 
analysis can reveal potential cancer risks, supporting early 
preventive strategies\cite{Wu2022}.

MedLMs also contribute to improving diagnostic processes. By 
analyzing imaging data, they assist in identifying abnormalities 
that might be missed in manual reviews. Some models trained on 
dermatological imaging have been shown to classify skin 
conditions with a high degree of accuracy, providing clinicians 
with additional tools for decision-making.

In treatment planning, MedLMs combine imaging, genomic, and 
historical patient data. This combination supports the 
development of personalized strategies for managing complex 
conditions. For diseases such as breast cancer, integrating 
these data sources has been associated with better-aligned 
treatment options for patients.

Drug discovery is another area where MedLMs show promise. They 
assist in predicting protein structures, which is a critical 
step in therapeutic development. Tools like AlphaFold have been 
used to reduce the time and effort required for molecular 
design, helping researchers identify drug candidates more 
efficiently\cite{Jumper2021}.

Medical large models (MedLMs) are experiencing widespread 
adoption across healthcare systems globally. Their utilization 
is increasing rapidly, particularly in areas such as diagnostic 
assistance, disease prediction, personalized treatment planning, 
and drug discovery. These models have proven their value in 
enhancing the accuracy and efficiency of medical practices. 
Notably, many healthcare platforms have now integrated these 
models as standard tools, embedding them deeply into their 
operations.

For instance, Baidu's Lingyi Medical Model leverages MedLM 
technology to enhance diagnostic accuracy, supporting doctors in 
making more precise disease predictions and diagnoses. By 
analyzing vast datasets, the model assists healthcare 
professionals in better understanding complex conditions, 
leading to improved treatment outcomes. Additionally, MedGPT, 
developed by Yilian, is another example of a medical language 
model that facilitates the entire healthcare process, from 
intelligent consultation to diagnosis recommendations and 
personalized treatment plans. This model integrates seamlessly 
into clinical workflows, helping doctors save time and make 
informed decisions based on comprehensive data analysis.

These platforms are just a few examples where MedLMs have become 
integral to healthcare technology. The incorporation of such 
models is not limited to the aforementioned systems; other 
companies, such as Yuanxin Technology and Jingtai Technology, 
are also employing these models to enhance patient management, 
accelerate drug development, and provide intelligent medical 
services. By embedding MedLMs as standard tools, these platforms 
are setting a new benchmark for healthcare innovation and 
improving overall patient care.

With these advancements, the high usage rate of MedLMs continues 
to rise, demonstrating their importance in modern healthcare 
settings. As the demand for precise and efficient healthcare 
grows, more medical institutions and research organizations are 
expected to adopt these models to support and enhance clinical 
decision-making, ultimately improving the quality of healthcare 
worldwide.

\begin{figure*}[!ht]
    \centering
    \includegraphics[width=0.98\textwidth]{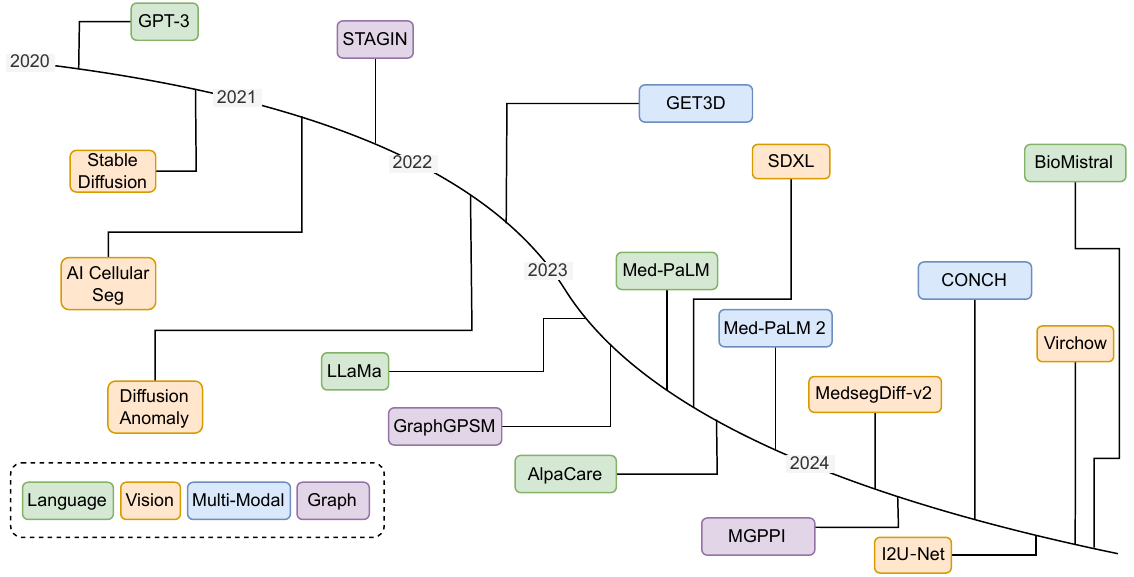}
    \vspace{-10pt}
    \caption{Evolution timeline of Large Models and their 
applications in medical.Including language models, vision 
models, multi-modal models and graph-based models.}
    \label{fig:timeline}
\end{figure*}

\section{LLMs in Medical}
\label{sec:formatting}

The effectiveness of large language models (LLMs) in medical 
question answering hinges on their training methodologies and 
the quality of the datasets used\cite{singhal2023large}. Leading 
research organizations, such as Google, have adopted advanced 
pretraining and fine-tuning methodologies to optimize the 
performance and efficacy of these models\cite{distilling2024}. 
For example, MedPaLM, built upon the general-purpose PaLM model, 
has been fine-tuned using high-quality, domain-specific medical 
datasets, enabling it to excel in understanding and reasoning 
about medical queries\cite{singhal2023expert}. To ensure the 
accuracy and comprehensiveness of these datasets, Google draws 
information from public medical databases (e.g., PubMed and 
MIMIC-III), professional guidelines, and detailed patient case 
records\cite{peng2019transfer}. These data sources are 
meticulously cleaned and reviewed by medical experts. 
Furthermore, multilingual processing capabilities are 
incorporated, enhancing the models' usability and robustness 
across diverse linguistic and cultural contexts\cite
{huang2024survey}. 

LLMs have rapidly become transformative tools within the medical 
field, offering significant advancements in medical 
education\cite{future2024}. For students preparing for the 
United States Medical Licensing Examination (USMLE), LLMs serve 
as invaluable resources. They excel in solving intricate 
clinical reasoning problems, simulating exam scenarios, and 
providing in-depth explanations for incorrect answers, which 
significantly enhances students' exam preparation and 
understanding of key concepts\cite{nori2023capabilities}. The 
success of LLMs in medical question-answering is largely 
attributed to the use of high-quality datasets. General-purpose 
datasets, such as PubMedQA and MIMIC-III, are essential for 
assessing clinical reasoning. While more specialized datasets 
like MedQA-USMLE, BioASQ, and MedMCQA address specific needs in 
the field1. For instance, MedQA-USMLE provides 
question-and-answer pairs that closely align with the structure 
and content of the USMLE\cite{medexqa2024}, while BioASQ focuses 
on biomedical knowledge retrieval and response generation\cite
{krithara2023bioasq}. By harnessing these diverse datasets, LLMs 
can generate precise, reliable, and contextually appropriate 
responses, solidifying their role as crucial tools in advancing 
medical education and practice.

AI large language models, such as GPT, are increasingly playing 
a pivotal role in the medical field, particularly in 
facilitating patient self-diagnosis and providing health 
management guidance \cite{shahsavar2023user}. For example, 
Pahola offers reliable, alcohol-related information to a global 
audience, thereby contributing to the effective implementation 
of Screening and Brief Interventions (SBI) and enhancing alcohol 
health literacy\cite{monteiro2022development}. Similarly, 
research has demonstrated that many individuals utilize ChatGPT 
for self-diagnosis and to access health information. An example 
of this is ChatGPT's utility in aiding patients in identifying 
common orthopedic conditions, such as carpal tunnel syndrome 
(CTS), prior to seeking consultation with healthcare 
professionals\cite{kuroiwa2023potential}. Moreover, AI-driven 
tools enable individuals to assess their mental health 
conveniently from the privacy of their homes\cite
{wimbarti2024critical}. 

To assess the performance of LLMs in medical question-answering 
tasks, researchers employ diverse and rigorous evaluation 
metrics. Standard metrics like accuracy, precision, recall, and 
F1 scores are used to measure the correctness and 
comprehensiveness of model responses\cite{llm2024}. For tasks 
requiring generated answers, BLEU scores evaluate the linguistic 
alignment between model outputs and reference answers\cite
{chen2019evaluating}. Moreover, domain-specific metrics, such as 
medical relevance and clinical applicability scores, are 
employed to gauge the professional and practical utility of the 
responses\cite{abbasian2024foundation}. Ethical and safety 
assessments are equally critical, utilizing measures such as 
harmful content detection rates and fairness evaluations to 
ensure the models' reliability and equity in medical 
applications\cite{farhud2021ethical}. These comprehensive 
evaluation frameworks support the continuous improvement and 
refinement of LLMs. Notably, existing LLMs may produce incorrect 
content or known as ``hallucinations''. More details will be 
discussed in later sections.


\section{Large Vision Models in Medical}
\subsection{Visual Anomaly Detection in Medical Images}

Medical image anomaly detection represents a critical component 
in computer-aided diagnosis systems, serving as an essential 
tool for early disease detection and treatment planning. Despite 
significant advances in deep learning, the inherent complexity 
of medical anomalies, coupled with the scarcity of annotated 
pathological data, continues to pose substantial challenges. As 
highlighted by the comprehensive BMAD benchmark \cite
{bao2024bmad}, which spans across five distinct medical domains, 
the field requires robust and generalizable approaches that can 
perform effectively across different modalities and anatomical 
structures. Recent developments in foundation models and 
generative approaches have introduced promising paradigms for 
addressing these challenges, revolutionizing how we approach 
medical anomaly detection.

\begin{table*}[!t]
	\centering
	\caption{Comparison of Different Approaches in Medical 
Visual Anomaly Detection}
	\label{tab:comparison}
	\begin{tabular}{p{2.3cm}|p{5.5cm}|p{4.5cm}|p{4cm}}
		\toprule
		\textbf{Strategy} & \textbf{Medical Applications} & 
\textbf{Evaluation Metrics} & \textbf{Key Findings} \\
		\midrule
		Vision-Language Models (Zero-shot) & 
		\begin{itemize}
			\item Brain tumor detection (MRI) \cite
{zhou2023anomalyclip}
			\item Chest abnormality detection (X-ray) \cite
{zhang2023biomedclip}
			\item Retinal disease screening \cite
{huang2024adapting}
		\end{itemize} & 
		\begin{itemize}
			\item AUROC \cite{park2024contrastive}
			\item Dice coefficient \cite{aleem2024test}
			\item Sensitivity/Specificity \cite
{koleilat2024medclip}
			\item Zero-shot transfer accuracy \cite
{zhang2023biomedclip}
		\end{itemize} & 
		\begin{itemize}
		\item Performance varies significantly across modalities 
\cite{huang2024adapting}	
		\item Superior in anatomical structure detection but 
struggles with subtle pathological changes \cite
{park2024contrastive} 
		\end{itemize}\\
		\midrule
		Diffusion Models & 
		\begin{itemize}
			\item Lesion detection in brain MRI \cite
{wolleb2022diffusion}
			\item Lung nodule detection in CT \cite
{fontanella2024diffusion}
			\item Histopathological analysis \cite
{behrendt2024diffusion}
		\end{itemize} & 
		\begin{itemize}
			\item FID score \cite{wolleb2022diffusion}
			\item SSIM \cite{behrendt2024diffusion}
			\item Image reconstruction error \cite
{iqbal2023unsupervised}
			\item Lesion detection rate \cite{fan2024discrepancy}
		\end{itemize} & 
		\begin{itemize}
		\item Excels in generating realistic counterfactuals 
\cite{fontanella2024diffusion}
		\item Computational overhead remains a challenge \cite
{wu2024medsegdiff} 
		\end{itemize}	\\
		\midrule
		Self-supervised Learning & 
		\begin{itemize}
			\item Multi-organ anomaly detection \cite
{tian2023self}
			\item Dermatological lesion analysis \cite
{iqbal2023unsupervised}
			\item Dental radiography \cite{wu2024medsegdiff}
		\end{itemize} & 
		\begin{itemize}
			\item Contrastive loss \cite{tian2023self}
			\item Reconstruction quality \cite
{iqbal2023unsupervised}
			\item Feature clustering metrics \cite
{wu2024medsegdiff}
			\item Anomaly detection accuracy \cite{tian2023self}
		\end{itemize} & 
		\begin{itemize}
			\item Effective with limited labeled data \cite
{iqbal2023unsupervised}
			\item Robust to domain shift but requires careful 
architecture design \cite{wu2024medsegdiff} 
		\end{itemize}	\\
		\midrule
		Semi-supervised Learning & 
		\begin{itemize}
			\item Pathological tissue classification \cite
{zhang2024spatial}
			\item Cardiac MRI analysis \cite{cai2023dual}
			\item Bone abnormality detection \cite
{ozbey2023unsupervised}
		\end{itemize} & 
		\begin{itemize}
			\item Classification accuracy \cite{zhang2024spatial}
			\item Segmentation IoU \cite{cai2023dual}
			\item Cohen's Kappa \cite{ozbey2023unsupervised}
			\item Expert consensus correlation \cite
{amit2023annotator}
		\end{itemize} & 
		\begin{itemize}
			\item Balanced performance between supervised and 
unsupervised approaches \cite{zhang2024spatial}; 
			\item Effective in clinical settings with partial 
annotations \cite{cai2023dual} 
		\end{itemize} \\
		\bottomrule
	\end{tabular}
\end{table*}

\subsubsection{Vision-Language Models for Zero-shot Medical 
Anomaly Detection}

The emergence of vision-language foundation models has marked a 
significant breakthrough in medical anomaly detection, 
particularly in zero-shot scenarios. These models, pre-trained 
on vast corpora of image-text pairs, offer a promising solution 
to the perennial challenge of limited labeled medical data. Zhou 
et al. \cite{zhou2023anomalyclip} introduced AnomalyCLIP, a 
pioneering approach that adapts CLIP (Contrastive Language-Image 
Pretraining) for zero-shot anomaly detection through 
object-agnostic prompt learning. The key innovation lies in 
learning text prompts that capture generic normality and 
abnormality patterns independent of foreground objects, enabling 
generalized anomaly recognition across diverse medical domains.

Building upon this foundation, Huang et al. \cite
{huang2024adapting} developed a sophisticated multi-level 
adaptation framework that significantly enhances CLIP's capacity 
for medical anomaly detection. Their approach incorporates 
multiple residual adapters into the pre-trained visual encoder, 
guided by pixel-wise visual-language feature alignment loss 
functions. This architecture effectively recalibrates the 
model's attention from general object semantics to specific 
medical anomaly patterns, achieving remarkable improvements in 
both anomaly classification and segmentation tasks.

The integration of SAM with CLIP has opened new avenues for 
zero-shot medical image analysis. Aleem et al. \cite
{aleem2024test} presented SaLIP, a cascade framework that 
combines SAM's precise segmentation capabilities with CLIP's 
semantic understanding. This unified approach demonstrates 
superior performance in organ segmentation tasks without 
requiring extensive domain-specific training data or manual 
prompt engineering. Furthermore, Koleilat et al. \cite
{koleilat2024medclip} extended this concept with MedCLIP-SAMv2, 
introducing a novel Decoupled Hard Negative Noise Contrastive 
Estimation loss and Multi-modal Information Bottleneck for 
enhanced segmentation performance.

The development of domain-specific models, exemplified by 
BiomedCLIP \cite{zhang2023biomedclip}, represents another 
significant advancement. Trained on an unprecedented 15 million 
biomedical image-text pairs, BiomedCLIP exhibits remarkable 
zero-shot capabilities across various medical imaging tasks, 
outperforming even specialized models in their respective 
domains. Park et al. \cite{park2024contrastive} further refined 
these approaches by introducing Contrastive Language Prompting 
(CLAP), specifically addressing the challenge of false positives 
in medical anomaly detection through careful prompt engineering.

\subsubsection{Diffusion Models for Unsupervised Detection}

Diffusion models have emerged as a powerful framework for 
unsupervised anomaly detection in medical imaging, offering 
unique advantages in modeling complex data distributions and 
generating high-quality counterfactuals. The pioneering work of 
Wolleb et al. \cite{wolleb2022diffusion} combined denoising 
diffusion implicit models with classifier guidance, 
demonstrating superior performance in preserving fine anatomical 
details compared to traditional GAN-based methods.

A significant advancement in this domain came from Fontanella et 
al. \cite{fontanella2024diffusion}, who introduced a novel 
approach for generating healthy counterfactuals of diseased 
images. Their method uniquely combines DDPM (Denoising Diffusion 
Probabilistic Models) and DDIM (Denoising Diffusion Implicit 
Models) at each sampling step, using DDPM to modify lesion areas 
while employing DDIM to preserve normal anatomy. This careful 
balance between modification and preservation has proven crucial 
for accurate anomaly detection.

The development of masked diffusion models represents another 
major innovation. Iqbal et al. \cite{iqbal2023unsupervised} 
introduced mDDPM, incorporating both Masked Image Modeling (MIM) 
and Masked Frequency Modeling (MFM) to enhance the model's 
ability to learn anatomically consistent representations. Liang 
et al. \cite{liang2023modality} extended this concept with their 
MMCCD framework, introducing cyclic modality translation as a 
mechanism for anomaly detection in multimodal MRI.

Recent work by Behrendt et al. \cite{behrendt2024diffusion} has 
focused on addressing the limitations of traditional anomaly 
scoring functions. Their approach introduces an adaptive 
ensembling strategy using Structural Similarity (SSIM) metrics, 
offering a more pathology-agnostic scoring mechanism that 
captures both intensity and structural disparities. Fan et al. 
\cite{fan2024discrepancy} further advanced this field with their 
discrepancy distribution medical diffusion (DDMD) model, which 
innovatively translates annotation inconsistencies into 
distribution discrepancies while preserving information within 
homogeneous samples.

\subsubsection{Self-supervised Learning for Anomaly Detection}

Self-supervised learning has emerged as a powerful paradigm for 
leveraging unlabeled medical data effectively. Tian et al. \cite
{tian2023self} introduced PMSACL, a groundbreaking 
self-supervised pre-training method that contrasts normal image 
classes against multiple pseudo classes of synthesized abnormal 
images. This approach addresses the critical challenge of 
learning effective low-dimensional representations capable of 
detecting unseen abnormal lesions of varying sizes and 
appearances. By enforcing dense clustering in the feature space, 
PMSACL significantly improves the sensitivity of anomaly 
detection across diverse medical imaging modalities.

Building on the success of masked modeling techniques, Iqbal et 
al. \cite{iqbal2023unsupervised} developed a novel 
self-supervised framework incorporating both spatial and 
frequency domain masking strategies. Their approach enables the 
model to learn more robust and anatomically-aware 
representations without requiring explicit annotations. This 
innovation has proven particularly effective in detecting subtle 
anatomical variations that might indicate pathological 
conditions.

The advancement of self-supervised learning has also led to 
improved understanding of normal anatomical variations, crucial 
for accurate anomaly detection. Wu et al. \cite
{wu2024medsegdiff} demonstrated how transformer-based 
architectures can be effectively combined with self-supervised 
learning objectives to capture long-range dependencies and 
structural relationships in medical images, leading to more 
reliable anomaly detection systems.

\subsubsection{Semi-supervised Learning Approaches}

Semi-supervised learning approaches have shown remarkable 
promise in leveraging both labeled and unlabeled data 
effectively for medical anomaly detection. Zhang et al. \cite
{zhang2024spatial} developed SAGAN, a sophisticated framework 
incorporating position encoding and attention mechanisms to 
accurately focus on abnormal regions while preserving normal 
structures. Their approach innovatively relaxes the cyclic 
consistency requirements typical in unpaired image-to-image 
translation, achieving superior performance in generating 
high-quality healthy images from unlabeled data.

Cai et al. \cite{cai2023dual} proposed a groundbreaking 
dual-distribution discrepancy framework that effectively 
leverages unlabeled images containing anomalies. Their approach 
introduces normative distribution and unknown distribution 
modules, with intra-discrepancy and inter-discrepancy measures 
serving as refined anomaly scores. This method has demonstrated 
significant improvements across various medical imaging 
modalities, including chest X-rays, brain MRIs, and retinal 
fundus images.

The integration of semi-supervised learning with traditional 
generative models has also shown promising results. Özbey et al. 
\cite{ozbey2023unsupervised} demonstrated how adversarial 
diffusion models could be effectively combined with 
semi-supervised learning strategies to improve image translation 
and anomaly detection performance. Their SynDiff framework 
showcases the potential of leveraging partially labeled datasets 
to enhance the fidelity and accuracy of generated medical images.

\subsubsection{Challenges and Future Directions}

Despite these advances, several critical challenges remain in 
medical anomaly detection. First, the balance between model 
complexity and clinical practicality continues to be a 
significant concern. While diffusion models offer superior 
performance, their computational requirements can be prohibitive 
in clinical settings, as noted by Wu et al. \cite
{wu2024medsegdiff}. Second, the integration of multiple expert 
annotations remains challenging, though recent work by Amit et 
al. \cite{amit2023annotator} on consensus prediction offers 
promising directions.

Looking forward, the field appears to be moving toward more 
efficient and interpretable approaches. The success of 
frequency-guided methods \cite{li2023zero} and structure-aware 
adaptations suggests that future developments may focus on 
incorporating domain-specific medical knowledge into foundation 
models. Additionally, the growing interest in self-supervised 
and semi-supervised approaches indicates a shift toward methods 
that can better utilize the vast amounts of unlabeled medical 
data available while maintaining the high standards of accuracy 
required in clinical applications.

\subsection{Applications in Pathological Images.}
\subsubsection{Image Segmentation with U-Net}
In pathological image analysis, the main approaches include 
pathological image segmentation, anomaly detection, and image 
generation\cite{pathology1,pathology2}.
The goal of pathological image segmentation is to divide 
different tissue or lesion regions within an image. Generally, 
it is the foundation of tasks including cancer tissue detection, 
lesion structure analysis, cell counting and so on. Models for 
pathological image segmentation often use convolutional neural 
network (CNN)-based architectures, particularly the U-Net model. 
U-Net Model is proposed by Olaf Ronneberger et al. in 2015\cite
{pathology3}, and it is a CNN architecture for biomedical image 
segmentation. It is widely used in tasks including cell boundary 
segmentation and the delineation of lesion areas. The U-Net 
consists of two main components: the encoder path and the 
decoder path\cite{pathology4}. The encoder extracts features by 
transforming raw pathological images into low-resolution, 
high-semantic feature representations using convolutional and 
pooling operations, and it also introduces non-linearity through 
functions such as ReLU\cite{pathology5}. The decoder restores 
image resolution by upsampling the encoded features to match the 
size of the input images, finally producing the segmentation 
output\cite{pathology6,pathology7}. The decoder path involves 
two key processes: deconvolution, which upsamples feature maps 
to restore resolution, and skip connections, which combine 
features from corresponding layers in the encoder and decoder 
keep spatial information lost in encoder. Meanwhile, 
convolutional operations integrate low-level details with 
high-level semantics to refine the segmentation results\cite
{pathology8}. In addition, there are several improved models 
based on U-Net, such as the Attention U-Net, which was proposed 
by Oktay et al. in 2018\cite{pathology9}. This model introduces 
attention modules to enhance the model's focus on critical 
regions. Attention modules are integrated into each skip 
connection to dynamically adjust the features transferred from 
the encoder to the decoder. These modules calculate attention 
weights for the input features and amplify features of important 
regions while suppressing features of irrelevant areas. Based on 
the classic encoder-decoder structure of U-Net, the attention 
modules are embedded at key nodes along the decoding path and 
improve the model's focus on significant regions while 
maintaining overall segmentation accuracy.
Image segmentation can be used in various pathological imaging 
tasks, like cell and nuclear segmentation. For example, when 
diagnosing corneal endothelial health states, U-Net-based CNN 
have achieved high accuracy in segmenting endothelial cells 
across images of varying cell sizes, and achieve precise 
measurement of cellular morphological parameters (AUROC 0.92, 
DICE 0.86)\cite{pathology10}.

Moreover, because cell segmentation is the first step in 
quantitative tissue imaging data analysis and the basis for 
single-cell analysis, it is important in identifying malignant 
tumors. The abnormal enlargement of nuclei in cancer cells leads 
to a significantly increased nucleus-to-cytoplasm ratio, which 
is one of the criteria for diagnosing malignant tumors\cite
{pathology11}. Therefore, image segmentation can also be applied 
to the dividing of malignant regions in pathological images. For 
example, in squamous cell carcinoma (SCC), researchers utilized 
a patch dataset extracted from 200 digitized tissue images of 84 
patients to train a U-Net-based segmentation model. The model 
achieved a segmentation AUC of 0.89 on the test set, and the 
average segmentation time is 72 seconds per image, which shows 
higher efficiency compared to traditional manual segmentation 
methods\cite{pathology12}.
\subsubsection{Image Generation with GANs}
Pathological image generation uses artificial intelligence (AI) 
to produce high-quality synthetic pathology images and solves 
challenges such as limited datasets and annotation difficulties 
by enlarging and balancing datasets. Therefore, it is generally 
used for upstream tasks\cite{pathology13}. Generative 
adversarial networks (GANs), introduced by Ian Goodfellow et al. 
in 2014\cite{pathology14}, are the primary methods for image 
generation. GANs consists of two neural network, including 
Generator and Discriminator, and train adversarially to produce 
realistic data , and it shows outstanding performance in the 
image generation\cite{pathology15}. The method optimizes the 
generator and discriminator alternately. The generator samples 
from the real dataset, and inputs the sampled noise into the 
generator, and produces synthetic samples that resemble real 
data; the discriminator takes both real and synthetic data as 
input and judges whether each sample comes from the real dataset 
or is generated by the generator. The discriminator is updated 
to improve its ability to distinguish between real and synthetic 
data, while the generator is updated to enhance its ability to 
generate synthetic data\cite{pathology16}. Through multiple 
iterations, the generator produces samples that are realistic 
enough to make it difficult for the discriminator to distinguish 
between real and synthetic data and then complete the task of 
generating pathological images. Moreover, StyleGAN, a GAN-based 
model, can also be used for generating pathological images\cite
{pathology17}. In traditional GANs, random noise with a Gaussian 
or uniform distribution is directly inputted into the generator 
to produce images. In contrast, StyleGAN introduces a 
style-mapping network, which maps noise Z to a new latent space 
W. The W is more expressive and easier to use for controlling 
specific features in the generated images, which helps adjust 
things like color, texture, or shape, making the images more 
realistic and diverse. Additionally, independent random noise is 
injected at each layer to generate random details in the 
images\cite{pathology18}. StyleGAN also uses progressive 
growth\cite{pathology19}, where the generator and discriminator 
initially handle low-resolution images during early training. 
The model gradually increases the resolution as training 
progresses until it achieves the target high resolution.
To further improve the resolution of generated images, a 
multi-scale conditional GAN method has been proposed. This model 
used a pyramid-like structure to progressively increase the 
resolution while generating high-resolution images and 
maintaining global consistency and detailed features of 
glandular structures. Through adversarial training, the 
generator captures the global layout of glands and the 
micro-textures of cell nuclei. Multi-layer discriminators ensure 
the authenticity and consistency of the generated image\cite
{pathology20}. 
In renal pathology image analysis, the morphological 
characteristics of glomeruli provide critical diagnostic and 
prognostic information. To improve diagnostic efficiency, an 
automated method based on CNNs was proposed. This method used 
GAN-based generative data augmentation to generate glomeruli 
pathological images with various morphologies and improve data 
diversity and model performance. The results showed that after 
applying generative data augmentation, the sensitivity of the 
classification model increased from 0.7077 to 0.7623, and 
specificity improved from 0.9316 to 0.9443\cite{pathology21}.
\subsubsection{Anomaly Detection in Pathological Images}
In addition to the pathological image segmentation and image 
generation above, there is also a downstream task, which is 
anomaly detection of pathological images. This involves using 
extracted image features to identify and localize abnormal 
images. In anomaly detection, there are generally two types of 
methods, which are supervised anomaly detection and unsupervised 
anomaly detection\cite{pathology22}. 

Supervised anomaly 
detection requires a dataset of annotated images including both 
normal and abnormal images. This kind of classification tasks 
could be performed using CNN-based model\cite{pathology23}. CNN 
models extract features through convolutional layers, reduce the 
resolution of feature maps using pooling layers, and map the 
extracted high-dimensional features to classification outputs 
using fully connected layers. The final classification 
probabilities for normal or abnormal states are generated using 
Softmax or Sigmoid functions in the output layer\cite
{pathology24}.

For example, studies have shown that research based on deep CNNs 
shows significant advantages in classifying skin lesions. Using 
a dataset containing about 130000 images and covering almost 
2000 diseases, CNNs achieved great performance comparable in two 
binary classification tasks. Besides, the CNN achieved an 
accuracy of 72.1\% in a three-class disease partitioning task 
and 55.4\% in a nine-class disease partitioning task\cite
{pathology25}.

In unsupervised or self-supervised anomaly detection, the 
primary methods include contrastive learning\cite{pathology26} 
and generative models\cite{pathology23,pathology27}. Contrastive 
learning involves extracting high-level feature representations 
from normal samples to construct a feature space. Abnormal 
samples, due to their different feature distributions, deviate 
from the feature space of normal samples. If the deviation is 
sufficiently large, they are identified as anomalies. Generative 
models are similar to GANs. They use an encoder to extract 
high-dimensional feature representations from input data, 
capturing complex patterns and learning features. Subsequently, 
a decoder reconstructs the input data from these 
high-dimensional features to recreate the original image. For 
normal samples, as they conform to the feature distribution 
learned by the model, the reconstruction error is low. However, 
for abnormal samples, the reconstruction error is high, leading 
to their identification as anomalies. 

Hui Liu et al. proposed a 
deep learning framework based on weakly supervised contrastive 
learning. This framework uses self-supervised pretraining on 
large-scale unlabeled patches from whole slide images (WSI) to 
extract highly informative pathological features. Combined with 
multitask learning, the framework successfully inferred breast 
cancer-related gene expression, molecular subtypes, and clinical 
outcomes. Experiments showed that this method achieved 
outstanding performance across multiple datasets, and the 
generated spatial heatmaps were highly consistent with 
pathologists' annotations and spatial transcriptomics data. This 
highlights its potential in linking genotypes with phenotypes 
and in the clinical applications of digital pathology\cite
{pathology28}. 

Besides, in related research, a generative model 
combining GANs and autoencoders was employed for anomaly 
detection. By training the model on normal tissue data, 
regularization and multi-scale contextual data were used to 
improve generalizability. This approach achieved efficient 
anomaly detection on the toxicologic histopathology (TOXPATH) 
dataset, with an AUC of 0.953\cite{pathology29}.

In general, pathological slide images can be used in different 
upstream and downstream tasks, such as classification and 
segmentation tasks. These include using image segmentation for 
cell counting\cite{pathology30},applying image generation in 
upstream tasks to perform pathological data augmentation\cite
{pathology31},using image segmentation and anomaly detection for 
cancer diagnosis or the identification of other lesion 
regions\cite{pathology32,pathology33},and conducting tasks such 
as tumor grading and progression prediction\cite{pathology34}.

In addition to the neural network models mentioned above, the 
increasing computational capacity has led to the emergence of 
large models. These large models show powerful feature learning 
abilities through pretraining on large-scale datasets. In recent 
years, the concept of foundation models\cite{pathology35} has 
emerged. 

These models not only have larger parameter sizes, with 
commonly used architectures in the field of pathology image 
analysis including Vision Transformers, CLIP, and Mask2Former 
ranging from hundreds of millions to billions of parameters, but 
also use high-resolution pathology images for pretraining on 
large-scale datasets.

Generally, the architectures of these foundation models are 
based on Transformer\cite{pathology36}. Initially it was 
developed for natural language processing, now it has been 
extended to the visual field. Through the self-attention 
mechanism, it effectively models both global and local features 
in pathology images and improves performance of tasks of 
segmentation and classification. While traditional CNNs rely on 
convolutional kernels to extract local features, Transformers 
rely entirely on self-attention to model relationships between 
all regions of an image and therefore capture the 
interdependencies of different tissue structures in pathology 
slides more effectively. Like neural network models, many 
Transformer-based models are developed to address tasks such as 
segmentation, classification diagnosis, and multimodality.

For image segmentation tasks, in addition to U-Net, 
Transformer-based segmentation large models, such as 
Mask2Former, have introduced multi-task segmentation frameworks 
and self-attention mechanisms to achieve pathology image 
segmentation. Jia-Chun Sheng et al. studied the 
Transformer-based Mask2Former segmentation model and evaluated 
its performance on pathology datasets. Mask2Former uses 
Transformer-based architecture. It used Swin Transformer as the 
encoder to extract multi-scale features. A pixel decoder 
integrated and processed features of different resolutions. The 
Transformer decoder applied masked attention to focus on 
foreground regions. This design improves its ability to segment 
small objects. By pretraining on natural images, Mask2Former 
adapts well to small pathology datasets. It achieved great 
performance comparable to or better than task-specific methods 
on the CRAG and GlaS datasets. It also significantly reduces the 
computational cost of handling high-resolution features \cite
{pathology37}.

In addition, Visual Transformer-based classification models have 
been increasingly introduced in pathology. Unlike traditional 
methods, these large models use the global modeling capabilities 
of Transformers to handle complex pathology images, like 
gigapixel whole slide images (WSIs). 

In pathology image 
classification, Hanwen Xu et al. proposed Prov-GigaPath, which 
extracts local features through a tile-level encoder and 
integrates global context features from WSIs using LongNet\cite
{pathology38}. By being pretrained on real-world datasets 
containing 1.3 billion tiles, the model effectively captures 
both local and global patterns using DINOv2 and a masked 
autoencoder (MAE). Prov-GigaPath achieved better performance in 
25 out of 26 pathology tasks, including a 23.5\% improvement in 
AUROC and a 66.4\% improvement in AUPRC for EGFR mutation 
prediction\cite{pathology39}. 

Eugene Vorontsov et al. proposed 
the Virchow model. It was based on Visual Transformer 
architecture and trained on 1.5 million WSIs. Through the DINOv2 
algorithm to learn both global and local embedding of WSIs, the 
model achieved cancer detection with ROC of 0.95 for nine common 
and seven rare cancers\cite{pathology40}. Richard J. Che's team 
proposed the general-purpose self-supervised pathology model 
UNI. It was pretrained on over 100,000 diagnostic H\&E-stained 
WSIs and showed exceptional performance in 34 computational 
pathology tasks, including cancer subtype generalization\cite
{pathology41}.

Recently, multimodal foundation models have gained attention in 
pathology image analysis. By integrating visual and textual 
information, these models simultaneously capture the structural 
features of pathological images and the semantic information of 
clinical text. 

CONCH is a foundational visual-language model 
that uses an image encoder, text encoder, and multimodal 
decoder. Through contrastive learning, it embeds images and text 
into a shared representation space while optimizing multimodal 
understanding through a captioning objective. During 
pretraining, CONCH leverages diverse pathology image-text pairs 
for unsupervised learning, significantly enhancing feature 
extraction capabilities. In cancer subtype classification, CONCH 
achieved 91.3\% zero-shot accuracy \cite{pathology42}.

The commonly used evaluation metrics in pathological image tasks 
are as follows:

\begin{table*}[htbp]
  \centering
  \caption{Evaluation Methods for Different Tasks}
  \begin{tabular}{>{\RaggedRight\arraybackslash}m{3cm} >
{\RaggedRight\arraybackslash}m{3.5cm} >
{\RaggedRight\arraybackslash}m{9cm} l l}
      \toprule
      \textbf{Task Type} & \textbf{Evaluation Method} & \textbf
{Description} & & \\
      \midrule
      \multirow{2}{3cm}[-2.5ex]{\centering Image Segmentation 
\cite{pathology43}} & Dice Coefficient & Measures the 
overlap between the predicted segmentation and the ground 
truth, ranging from [0, 1], with higher values indicating 
better performance. & & \\
      \cmidrule{2-5}
      & IoU (Intersection over Union) & Measures the ratio of 
the intersection area to the union area between the 
predicted and ground truth segmentation regions, ranging 
from [0, 1]. & & \\
      \midrule
      \multirow{3}{3cm}[-8.5ex]{Image Generation \cite
{pathology44,pathology45,pathology46}} & FID (Fréchet 
Inception Distance) & Assesses the difference in feature 
distributions between generated and real images; lower 
values indicate better performance. & & \\
      \cmidrule{2-5}
      & IS (Inception Score) & Evaluates the diversity and 
quality of generated images based on a classification 
model; higher values indicate better performance. & & \\
      \cmidrule{2-5}
      & Evaluation Methods in Downstream Tasks & The evaluation 
of image generation considers not only image quality (e.
g., FID, IS) but also its impact on downstream tasks like 
classification and segmentation, using metrics such as 
accuracy or sensitivity to validate effectiveness. & & \\
      \midrule
      \multirow{5}{3cm}[-8.5ex]{Anomaly detection} & Accuracy & 
Proportion of correctly classified samples, reflecting 
overall classification performance. & & \\
      \cmidrule{2-5}
      & Sensitivity/Recall & Ability to detect anomalous samples 
(recall); higher values indicate better performance. & & \\
      \cmidrule{2-5}
      & Specificity & Ability to correctly detect normal 
samples; higher values indicate better performance. & & \\
      \cmidrule{2-5}
      & F1 Score & Harmonic mean of precision and recall, 
providing a balanced performance metric, ranging from [0, 
1]. & & \\
      \cmidrule{2-5}
      & ROC & Plots the curve of sensitivity against the false 
positive rate; performance is measured by AUC, with values 
closer to 1 indicating better performance. & & \\
      \bottomrule
  \end{tabular}
\end{table*}

\section{Applications of Large Multimodal Model in Medical}

Vision language models (VLMs) are multimodal generative AI 
models capable of reasoning over text, image, and video prompts. 
VLM has demonstrated excellent processing capabilities in image 
\& video generation, text-centric visual question answering, 
zero-shot detection, video summarization, and other fields, and 
some of the results have been successfully applied to autonomous 
driving, AI-generated image \& text generation, AR/VR techniques 
and other fields \cite{Vlm3D1}. In the past, machine 
learning-based graphics processing technology has been widely 
used for the diagnostics of radiology and medical education \cite
{Vlm3D2,Vlm3D3}. As a new generation of image processing model, 
VLM has attracted the attention of medical workers with its 
potential, and its application as the superior substitution of 
traditional image processing models in the medical field is 
becoming another hot research direction.

In computer graphics, 3D modeling refers to the process of 
developing mathematical coordinate-based representations of the 
surface of objects (inanimate or biological) in three dimensions 
by manipulating edges, vertices and polygons in simulated 3D 
space through specialized software. 3D modeling has been widely 
used in architecture and industrial design, video games, film 
industry, art and other fields, but traditional 3D modeling 
technique has a strong dependence on human resources and time 
cost. As a result, the cost and quality of the final product 
could be difficult to control. The recent emergence of 3D large 
models is bringing new changes to this field \cite{Vlm3D4}. 
Three-dimensional modeling was applied to multiple types of 
tasks in medicine (such as prosthetic and implant design, image 
reconstruction, anatomical modeling, medical education, etc.), 
and 3D large models may bring a technical revolution to these 
applications.

\subsection{VLM for Medical Image Analysis}
The models and algorithms based on traditional machine learning 
models (such as CNN) that have been widely used in the research 
of medical imaging \cite{Vlm3D5}. VLM as a technique that has 
shown great capabilities in processing complex relationships and 
multimodal information \cite{Vlm3D1}, is naturally adapted to 
the complexity of information contained in medical cases (text, 
charts, pictures, videos, audio, etc.); therefore, VLM is 
gradually attracting the attention of medical workers.

The application of VLM large models in medicine at present is 
mainly focused on the processing of medical images, especially 
for pathological and radiological imaging, for which have a much 
greater demand for softwires with efficient graphic processing 
capabilities than other subspecialties. There are already some 
studies involving VLM large models for radiology and pathology, 
and this topic is discussed in details at other articles of our 
journal.

\subsection{VLM for Medical Single Analysis}
VLM models are also applied to the analysis of physiological 
signals. Just like traditional machine learning models, 
researches about the availability of VLMs for ECG, EEG analysis 
are also being conducted \cite{Vlm3D9,Vlm3D10,Vlm3D11}.

Due to the high correlation between VLM and computer vision, VLM 
is currently also used in the field of AI-based robotics. 
Currently, some EEG and sEMG combined technologies have been 
experimentally applied to the collection of limb movement data 
and the optimization of prosthetic movement patterns \cite
{Vlm3D12,Vlm3D13,Vlm3D14}. Meta has already released two large 
datasets and benchmarks for sEMG-based typing and pose 
estimation in Dec. 2024 \cite{Vlm3D15}.

\subsection{3D Large Models for Biometric Analysis}
3D modeling technology has been widely used in the medical 
field, especially in human anatomical modeling, prosthetic 
design and 2D to 3D image conversion \& construction \cite
{Vlm3D6,Vlm3D7,Vlm3D8}. With the advent of 3D large models, this 
technology has gradually been applied to the research in 
corresponding medical fields.

Another area of 3D modeling--3D human structure construction, 
has also been influenced by 3D large model with the production 
of some new research progresses. 3D model construction has been 
applied for educational use to better demonstrate body 
structures. In addition, 3D printing is widely used in 
orthopedics, surgery, instrument design, drug research and many 
others in recent years.

\subsection{Discussion and Outlook}
Diagnostic techniques in neurology, psychiatry based on machine 
learning algorithms have been one of the hot areas in AI related 
medical research in the past few years. Motion capture in 
combination with deep learning is also applied to the research 
of clinical measurements for physical medicine and 
rehabilitation \cite{Vlm3D16,Vlm3D17}. There are some studies 
proved to be useful on the screening \& diagnosis of stroke, 
Parkinson's disease as well as some mental disorders based on 
facial expression analysis \cite{Vlm3D18,Vlm3D19,Vlm3D20}. VLM, 
as a new generation of graphic image model, also have strong 
prospects in these directions. At present, some results of 
emotion analysis research related to facial expression capture 
based on VLM have been published \cite{Vlm3D21,Vlm3D22,Vlm3D23}. 
As algorithms develop and mature in the future, we can foresee 
that VLM will become more and more important in screening and 
remote diagnosis.
\begin{table*}[h]
	\centering
	\caption{Evaluation Methods for Different Models}
	\begin{tabular}{>{\RaggedRight\arraybackslash}m{3cm} >
{\RaggedRight\arraybackslash}m{5cm} >
{\RaggedRight\arraybackslash}m{7cm} l l}
		\toprule
		\textbf{Model Type} & \textbf{Application} & \textbf
{Evaluation Metrics} & & \\
		\midrule
		\multirow{3}{3cm}[-20ex]{\centering VLMs} & Medical 
Image Segmentation & \begin{tabular}[c]{@{}l@{}}Dice 
score \cite{Vlm3D32}\\ mIoU metric \cite{Vlm3D32}\end
{tabular} & & \\
		\cmidrule{2-5}
		& {\vspace{15ex} Medical report generation} & \begin
{tabular}[c]{@{}l@{}}Bilingual Evaluation Understudy/
BLEU \cite{Vlm3D33}\\ Recall-Oriented Understudy for 
Gisting \\ Evaluation/ROUGE \cite{Vlm3D33}\\ Metric for 
Evaluation of Translation with \\ Explicit ORrdering/ 
METEOR \cite{Vlm3D33}\\ Perplexity\cite{Vlm3D33}\\ 
BERTScore \cite{Vlm3D33}\\ RadGraph F1 \cite{Vlm3D33}\\ 
Human evaluation \cite{Vlm3D33}\\ Clinical efficacy 
metrics\\ Accuracy \cite{Vlm3D33}\\ Precision \cite
{Vlm3D33}\\ Recall \cite{Vlm3D33}\\ F1 \cite{Vlm3D33}\end
{tabular} & & \\
		\cmidrule{2-5}
		& Question answering & \begin{tabular}[c]{@{}l@{}}
Accuracy \cite{Vlm3D33}\\ Exact match \cite{Vlm3D33}\\ 
Human evaluation \cite{Vlm3D33}\end{tabular} & & \\
		\midrule
		\multirow{3}{3cm}[-6.5ex]{3D large models} & 3D Medical 
Image Analysis & \begin{tabular}[c]{@{}l@{}}Dice score 
\cite{Vlm3D34}\\ Hausdorff Distance 95\%/HD95 \cite
{Vlm3D34}\end{tabular} & & \\
		\cmidrule{2-5}
		& Medical Image Segmentation \& generation & Dice score 
\cite{Vlm3D35} & & \\
		 \cmidrule{2-5}
		& High-quality Medical Image Generation & \begin{tabular}
[c]{@{}l@{}}Frechet Inception Distance/FID \cite{Vlm3D36}
\\ Maximum Mean Discrepancy \cite{Vlm3D36}\\ Peak 
Signal-to-Noise Ratio/PSNR \cite{Vlm3D36}\\ Structural 
Similarity Index/SSIM \cite{Vlm3D36}\end{tabular} & & \\
		\bottomrule
	\end{tabular}
  \end{table*}
  The audio and video generation function based on VLM and 3D 
large models can also more vividly show the required content 
to medical workers and patients, so as to be applied for 
demonstration, diagnosis explanation, education and 
popularization. Medical knowledge based on 3D models has also 
been proven to have more advantages than traditional forms 
\cite{Vlm3D24,Vlm3D25}. 3D large models can optimize 
traditional technology-based 3D anatomical structure 
construction and apply it to medical education and 
popularization. Traditional 3D model teaching software, such 
as visual body, has a certain distance from the requirements 
of actual teaching tasks in terms of the richness and accuracy 
of details. Many teaching scenes or structures that need to be 
understood in clinical applications are often not recorded or 
lack proper accuracy in these softwires. 3D large models can 
not only provide better contents, but also customize them 
based on user needs (such as text image conversion, model 
scaling, highlighting, separation, etc.) \cite{Vlm3D4}. In 
addition, 3D large models can also be applied to 3D printing 
to produce realistic human anatomy models \cite{Vlm3D26}, 
which may better serve the purpose of demonstration.

  In the past few decades, the development of prosthetics 
related to machine vision has always attracted many experts. 
The relevant mechanical design has long been mature. However, 
due to algorithm, materials along with other technical 
obstacles, traditional prosthetic designs always have 
limitations in final functions and user experience. The recent 
introduction of multimodal LLM and neuro-physical 
interface-related technologies has brought changes to this 
field. Auxiliary algorithms based on VLM and 3D large models, 
machine vision and image analysis technologies may become the 
key puzzle to solve the problem. Now with the help of AI 
techniques, there's finally a chance of modern prosthetics to 
achieve a difference from hooks and sticks.
  The researchers at Massachusetts Institute of Technology 
recently published a study on intelligent prosthetic design 
based on mechanism of antagonistic muscle groups \cite
{Vlm3D27}. It can be seen that 3D large model technology can 
provide considerable assistance to the design of implants and 
the surgical implantation based on the construction and 
analysis of anatomical structures.
  
  Machine vision and image analysis based on VLM and 3D large 
models can also be applied to the research of personalized 
intelligent vehicles, thereby providing a better experience 
for paralyzed patients or those who rely on wheelchairs. Take 
Simultaneous Localization and Mapping/SLAM as an example, it 
is a technique with remarkable capabilities on construct 
surrounding environments. SLAM system can provide reference 
information for the action strategies, which makes it a key 
component in medical robotic design. As SLAM has its own 
limitations on map semantically similar objects in compact 
environments \cite{Vlm3D28}, VLM's capabilities on image 
analysis make it a choice with great potential to guide the 
actions of robotics in SLAM constructed models, which can 
further be applied for aforementioned fields.
  
  In the future, 3D large models would be trained with a larger 
amount of data, which may facilitate to establish a unified 
database in the corresponding field, and construct subject 
models with a certain degree of universality, thereby saving 
the resources and time cost for personalized adjustment of 
modeling. 3D modeling has already been used to aid procedures 
in endoscopy and some surgeries, it is also applied to aid the 
production of implants \cite{Vlm3D29,Vlm3D30}. The 
introduction of 3D large models may lower the overall cost of 
prosthetics \& implants customization for general surgery, 
orthopedics, plastic surgery, dentistry and other disciplines, 
and further guide the progress in these fields.
  
  Micro-robots and soft robots have always been one of the hot 
areas of translational medicine. Research results in the field 
of soft robotics are gradually making such technologies 
practical. In these researches, 3D large models can provide 
basic support for anatomical structure reconstruction and 
dynamical simulation of such designs. 
  
  \subsection{Limitations}
  Medical imaging will not be affected by image editing or false 
information like other contents, but the accuracy of trained 
models based on the complexity of the disease is an important 
issue.
  Take neurology as an example. In brain CT and MRI of difficult 
cases, the tiny lesions inside the brain often can only be 
visualized under the scale of millimeter level with strong 
interferences, and are also greatly affected by the equipment, 
shooting angle, patient compliance along with many other 
issues. For example, in general neurological imaging, a 
trained radiologist sometimes cannot correctly find the 
lesion, so the participation of superior specialists from 
neuroimaging is often required. This usually involves a 
complete review and detailed analysis of the patient's medical 
history, and the accumulation of this part of knowledge and 
experience cannot be completely replaced by the current model 
in a short time. Although data training of large models will 
gradually make progress, before achieving more ideal results, 
the application of such research in clinical practice will 
inevitably be limited by the concerns of patient safety 
issues. In contrast, for popular science and patient 
education, the requirements for accuracy of image analysis and 
model construction are relatively loose compared to clinical 
applications and professional teaching, and may usher in 
mature transformation one step earlier.
  
  Neuro-physical interface is indeed a hot field at present, its 
combination with VLM and 3D large models is also very 
attractive. However, key technical issues related to BCI/NPI 
still exist \cite{Vlm3D31}. Limited by the development of 
materials science and computer science, even if the relevant 
technical conditions are complete, the patient's rejection 
reaction, the accuracy of signal measurement, the propagation 
delay transduction, the accuracy of motion conversion, the 
patient's tolerance of invasive \& non-invasive devices 
(appearance, size, weight, etc.) are all important issues 
related to the transformation of results.

\section{Large Graph Models in Medical}
\subsection{Introduction of Large Graph Models}
Large Graph Models (LGMs) are characterized by having a vast 
number of parameters, which endows it with significantly greater 
capabilities compared to smaller models\cite
{zhang2023graphmeetsllmslarge}. This enhanced capacity 
facilitates the understanding, analysis, and processing of tasks 
related to graph data. LGMs, particularly Graph neural networks 
(GNNs) and Graph transformer architectures, have emerged as 
powerful tools for processing and analyzing structured data, 
especially for complex and large-scale datasets. Their ability 
to capture complex relationships between entities makes them 
especially suitable for various applications in the medical 
field, where understanding such interactions is often crucial. 
LGMs utilizing transformer architecture and can be scaling up 
like LLMs. So they may requiring substantial computational 
overheads.

Healthcare primarily focus on management for patient care and 
administrative purposes rather than for data-driven research. 
Electronic health records (EHRs) typically consist of a mix of 
structured, semi-structured, and unstructured data, encompassing 
structured tables, images, waveforms, and clinical notes. These 
datasets capture a wide array of complex and interrelated 
concepts, leading to data that is inherently high-dimensional 
and heterogeneous\cite{data2016challenges} with sub-optimal data 
quality due to the rapid pace of the clinical environment and 
the absence of manual curation. 

Graphs offer a structured approach to explicitly model 
relational structures within data representations. Besides, 
graphs strike a balance between flexibility and structure when 
representing data, which allows for the seamless integration of 
multiple data modalities and the ability to exploit 
interdependencies across these modalities. This capability 
facilitates the mathematical incorporation of domain-specific 
prior knowledge to inform and enhance patient 
representations\cite{Nelson540963}. Graph AI streamlines the 
transfer of LGMs across various clinical tasks, allowing them to 
effectively apply to different patient populations with minimal 
or no additional parameters or retraining \cite
{johnson2023graphaimedicine}. 

GNNs are a type of neural network model specifically designed 
for graph-structured data. They can learn representations for 
nodes, edges, and entire graphs. The field of graph 
representation learning has developed a variety of network 
architectures, each tailored to capture distinct types of 
complex relationships within graph-structured data. For tasks 
involving predictions at the node, edge, or graph level, 
message-passing and transformer-based architectures are among 
the most prevalent\cite{zhang2023graphmeetsllmslarge}.

There are numerous applications for Large Graph Models in 
biomedical field, including Brain Network Analysis, Medical 
Knowledge Graphs, Drug Discovery Analysis, and Protein-Protein 
Interaction Networks.

preamble:
\begin{table*}
    \centering
	\caption{Summary of Large Graph Models with Datasets and 
Evaluation Metrics in the Medical Field. 
	\textcolor{gray}{Abbr.
	HCP (Human Connectome Project);
	STAGIN (Spatio-Temporal Attention Graph Isomorphism Network);
	ADNI (Alzheimer's Disease Neuroimaging Initiative);
	UMLS (Unified Medical Language System);
	GDSC (Genomics of Drug Sensitivity in Cancer);
	GNBR (Global Network of Biomedical Relationships);
	PDB (Protein Data Bank);
	HPRD (Human Protein Reference Database);
	OPHID (Online Predicted Human Interaction Database);
	BioGRID (H. sapiens dataset from the Biological General 
Repository for Interaction Datasets);
	AUC-ROC (Area Under the Curve - Receiver Operating 
Characteristic)}
	}
    \begin{tabularx}{\textwidth}{|>{\raggedright\arraybackslash}
X|>{\raggedright\arraybackslash}X|>
{\raggedright\arraybackslash}X|>{\raggedright\arraybackslash}
X|} 
    \hline
    Application & Datasets & Method/ Model & Evaluation Metrics 
\\ 
    \hline
    Brain Network Analysis & HCP-Rest,~ HCP-Task & STAGIN\cite
{NEURIPS2021_22785dd2} & Prediction accuracy, AUC-ROC \\ 
    \cline{2-4}
    & ADNI & LG-GNN\cite{9936686} & Prediction accuracy, 
Sensitivity, Specificity, F1 score, AUC-ROC \\ 
    \hline
    Medical Knowledge Graphs & Sampling diseases from the 
Knowledge Graphs & TxGNN (GNN)\cite{Huang2023_03_19_23287458} & Prediction accuracy \\ 
    \cline{2-4}
    & UMLS & DR.KNOWS\cite
{gao2023leveragingmedicalknowledgegraph} & Precision, 
Recall, F- score \\ 
    \hline
    Drug Discovery Analysis & Davis, Kiba & GraphDTA (GNN)\cite
{10_1093_bioinformatics_btaa921} & Prediction accuracy \\ 
    \cline{2-4}
    & GNBR & Graph embedding\cite{Sosa727925} & AUC-ROC \\ 
    \hline
    Protein-Protein Interaction Networks & AlphaFold Protein 
Structure Database, PDB & GraphGPSM (GNN)\cite{10_3389_fgene_2024_1440448} & TM-scores,~ Prediction accuracy \\ 
    \cline{2-4}
    & HPRD,~ OPHID,~ BioGRID,~ STRING & MGPPI (GNN)\cite{10_1093_bib_bbad219} & Prediction accuracy, Precision, Recall, F1 
score, AUC-ROC \\
    \hline
    \end{tabularx}
\end{table*}
\subsection{Brain Network Analysis}
Within the realm of neuroscience, GNNs are instrumental in 
analyzing complex datasets like brain connectomes. By modeling 
the brain as a network of interconnected regions, GNNs can 
identify patterns associated with neurological disorders. 

Graph-based approaches have offered valuable insights into brain 
functions by analyzing the connectome as a network, calculating 
functional connectivity (FC) between brain regions using 
functional neuroimaging. There are challenges remain in 
capturing the dynamic nature of FC networks, which fluctuate 
over time. Addressing these limitations, researchers built 
STAGIN\cite{NEURIPS2021_22785dd2}, a method that incorporates 
spatio-temporal attention to learn the dynamic graph 
representation of brain connectomes, integrating temporal 
sequences of brain graphs with READOUT functions and a 
Transformer encoder for spatial and temporal explainability. 
Experiments on HCP-Rest and HCP-Task datasets show superior 
performance of STAGIN, with its spatio-temporal attention 
mechanisms providing interpretations aligned with existing 
neuroscientific knowledge, validating the approach.

Functional brain networks have been increasingly utilized for 
classifying brain disorders such as Autism Spectrum Disorder 
(ASD) and Alzheimer's Disease (AD). Traditional approaches often 
overlook non-imaging information and inter-subject relationships 
or fail to identify disease-specific brain regions and 
biomarkers, resulting in less accurate classifications. To 
overcome these challenges, researchers introduce the 
local-to-global graph neural network (LG-GNN)\cite{9936686}, 
which includes a local ROI-GNN for extracting feature embeddings 
of brain regions and identifying biomarkers, and a global 
Subject-GNN that leverages these embeddings and non-imaging data 
to learn inter-subject relationships. The LG-GNN was validated 
on public datasets for ASD and AD classification, achieving 
state-of-the-art performance across various evaluation metrics.

\subsection{Medical Knowledge Graphs}
GNNs also enhance medical knowledge graphs, which aggregate vast 
medical knowledge to inform diagnostics and treatment plans. For 
example, drug repurposing is typically an opportunistic effort 
to find new uses for approved drugs. However, it faces 
limitations due to AI models' focus on diseases with existing 
treatments. To solve this problem, TxGNN is introduced as a 
graph foundation model for zero-shot drug repurposing, aiming to 
identify therapeutic candidates for diseases with few or no 
treatments. By leveraging a medical knowledge graph, TxGNN uses 
a graph neural network and metric learning to rank drugs for 
potential indications and contraindications across 17,080 
diseases. As a result, it outperforms eight other methods with 
improved prediction accuracy for both indications and 
contraindications\cite{Huang2023_03_19_23287458}. 

In addition, EHRs are crucial for comprehensive patient care, 
although their complexity and verbosity can overwhelm healthcare 
providers and lead to diagnostic errors. While Large Language 
Models (LLMs) show promise in various language-related tasks, 
their application in healthcare must prioritize accuracy and 
safety to prevent patient harm. An approach has been proposed 
that enhances LLMs in automated diagnosis generation by 
integrating a medical knowledge graph (KG) derived from the 
Unified Medical Language System (UMLS) and a novel graph model 
inspired by clinical diagnostic reasoning. Experiments with 
real-world hospital data reveal that this combined approach 
improves diagnostic accuracy and provides an explainable 
diagnostic pathway, advancing AI-augmented decision support 
systems in healthcare\cite
{gao2023leveragingmedicalknowledgegraph}.

\subsection{Drug Discovery Analysis} 
The process of discovering new therapeutic applications for 
existing drugs involves key challenges, such as modeling the 
intricate interactions among genes, pathways, targets, and 
drugs, which leads to an exponentially vast search space. 

GNNs are revolutionizing the process of drug discovery by 
predicting molecular properties essential for pharmaceutical 
efficacy. They provide insights into molecular interactions by 
modeling compounds as graphs. GraphDTA\cite{10_1093_bioinformatics_btaa921} is a new model which represents drugs as 
graphs and employs GNNs to predict drug-target affinity. It is 
found that GNNs not only achieve superior predictions of 
drug-target affinity compared to traditional non-deep learning 
models but also outperform other deep learning approaches. These 
results validate the suitability of deep learning models for 
predicting drug-target binding affinity and highlight the 
advantages of using graph representations for drugs in enhancing 
prediction accuracy.

Drug repurposing emerges as a promising alternative method for 
rare diseases, utilizing existing FDA-approved drugs for new 
therapeutic indications. To systematically generate drug 
repurposing hypotheses, it is crucial to integrate data from 
pharmacology, genetics, and pathology, which is facilitated by 
the Global Network of Biomedical Relationships (GNBR), a 
comprehensive knowledge graph. By applying a knowledge graph 
embedding method that models uncertainty and uses link 
prediction, this approach effectively generates and validates 
new drug repurposing hypotheses, achieving high performance 
(AUC-ROC = 0.89) and providing explanations for its 
predictions\cite{Sosa727925}.

\subsection{Protein-Protein Interaction Networks}
Protein structure scoring models are typically categorized into 
unified field and protein-specific functions, yet current 
prediction methods still fall short in accurately modeling 
complex structures, such as multi-domain and orphan proteins. In 
understanding disease mechanisms, GNNs help predict 
protein-protein interactions, which are vital for biological 
processes and pathways. 

Despite the preference for computational PPI prediction methods 
due to their cost-effectiveness and accuracy, many current 
approaches fall short in extracting detailed structural 
information and lack interpretability. MGPPI\cite{10_3389_fgene_2024_1440448}, a novel multiscale graph convolutional neural 
network, addresses these issues by effectively capturing both 
local and global protein structures and enhancing 
interpretability through Gradient Weighted Interaction 
Activation Mapping (Grad-WAM). Demonstrating superior 
performance across various datasets, MGPPI not only identifies 
key binding sites, such as those between the SARS-CoV-2 spike 
protein and human ACE2, but also highlights residues that could 
serve as biomarkers for predicting cancer patient survival, 
showcasing its potential for guiding personalized treatment and 
drug target identification.

In addition, GraphGPSM\cite{10_1093_bib_bbad219} is a new global 
scoring model based on an equivariant graph neural network 
(EGNN), has been developed to improve protein structure 
prediction and ranking. By integrating advanced features like 
residue-level ultrafast shape recognition and Gaussian radial 
basis function encoding with Rosetta energy terms, GraphGPSM 
shows a strong correlation with TM-scores on CASP13, CASP14, and 
CAMEO test sets, outperforming existing models like REF2015, 
ModFOLD8, and AlphaFold2, particularly in modeling challenging 
proteins.

\subsection{Conclusion}
Large Graph Models, with their profound ability to model complex 
data structures, have unprecedented potential in various medical 
applications, offering innovative solutions to complex 
challenges encountered in healthcare. Nonetheless, the 
significance of human-centered design and model interpretability 
in clinical decision-making remains paramount. As graph AI 
models derive insights through localized neural transformations 
on relational datasets, they present both opportunities and 
challenges in explaining model logic. Knowledge graphs can 
improve interpretability by aligning insights generated by the 
models with established medical knowledge. New graph AI models 
are emerging that integrate diverse data modalities through 
pre-training, facilitate interactive feedback loops, and 
encourage human-AI collaboration, ultimately leading to 
clinically relevant predictions\cite{johnson2023graphaimedicine}.

The future of GNNs in medicine hinges on continuing research and 
extensive datasets to refine these models further, 
interdisciplinary collaboration, driving forward the development 
of precision medicine and advancing the global health landscape.

{
    \small
    \bibliographystyle{ieeenat_fullname}
    \bibliography{main}
}


\end{document}